\documentclass[sigconf]{aamas} 

\usepackage{algorithm}
\usepackage{algorithmic}

\usepackage{comment}
\usepackage{amsmath}
\usepackage{makecell}
\usepackage{mathtools}
\usepackage{amsthm}
\usepackage{bbm}
\usepackage{enumitem}

\usepackage{amsfonts}
\usepackage{bm}
\usepackage{braket}
\usepackage{graphicx}
\usepackage[title]{appendix}
\usepackage{subcaption}

\DeclareMathOperator*{\esssup}{ess\,\!sup}
\DeclareMathOperator*{\argmax}{arg\,\!max}

\theoremstyle{definition}

\newtheorem{proposition}{Proposition}
\newtheorem{lemma}{Lemma}

\usepackage{balance} 

\setcopyright{ifaamas}
\acmConference[AAMAS '23]{Proc.\@ of the 22nd International Conference
on Autonomous Agents and Multiagent Systems (AAMAS 2023)}{May 29 -- June 2, 2023}
{London, United Kingdom}{A.~Ricci, W.~Yeoh, N.~Agmon, B.~An (eds.)}
\copyrightyear{2023}
\acmYear{2023}
\acmDOI{}
\acmPrice{}
\acmISBN{}

\acmSubmissionID{615}

\title{Toward Risk-based Optimistic Exploration\\ for Cooperative Multi-Agent Reinforcement Learning}

\author{Jihwan Oh$^{*}$}
\thanks{*Equal Contribution.}
\affiliation{
  \institution{Department of Economics and Law\\ Korea Military Academy}
  \city{Seoul}
  \country{South Korea}}
  \email{ericoh92920@gmail.com}

\author{Joonkee Kim$^{*}$}
\affiliation{
  \institution{Kim Jaechul Graduate School of AI\\ KAIST}
  \city{Seoul}
  \country{South Korea}}
\email{joonkeekim@kaist.ac.kr}

\author{Minchan Jeong}
\affiliation{
  \institution{Kim Jaechul Graduate School of AI\\ KAIST}
  \city{Seoul}
  \country{South Korea}}
\email{mcjeong@kaist.ac.kr}

\author{Se-Young Yun}
\affiliation{
  \institution{Kim Jaechul Graduate School of AI\\ KAIST}
  \city{Seoul}
  \country{South Korea}}
\email{yunseyoung@kaist.ac.kr}

\begin{abstract}
The multi-agent setting is intricate and unpredictable since the behaviors of multiple agents influence one another. To address this environmental uncertainty, distributional reinforcement learning algorithms that incorporate uncertainty via distributional output have been integrated with multi-agent reinforcement learning (MARL) methods, achieving state-of-the-art performance. However, distributional MARL algorithms still rely on the traditional $\epsilon$-greedy, which does not take cooperative strategy into account. In this paper, we present a risk-based exploration that leads to collaboratively optimistic behavior by shifting the sampling region of distribution. Initially, we take expectations from the upper quantiles of state-action values for exploration, which are optimistic actions, and gradually shift the sampling region of quantiles to the full distribution for exploitation. By ensuring that each agent is exposed to the same level of risk, we can force them to take cooperatively optimistic actions. Our method shows remarkable performance in multi-agent settings requiring cooperative exploration based on quantile regression appropriately controlling the level of risk.
\end{abstract}


\newcommand{\BibTeX}{\rm B\kern-.05em{\sc i\kern-.025em b}\kern-.08em\TeX}

\begin{document}


\pagestyle{fancy}
\fancyhead{}


\maketitle 


\section{Introduction}
\label{sec:introduction}

\begin{figure}[!t]
  \centering
  \includegraphics[width=\linewidth]{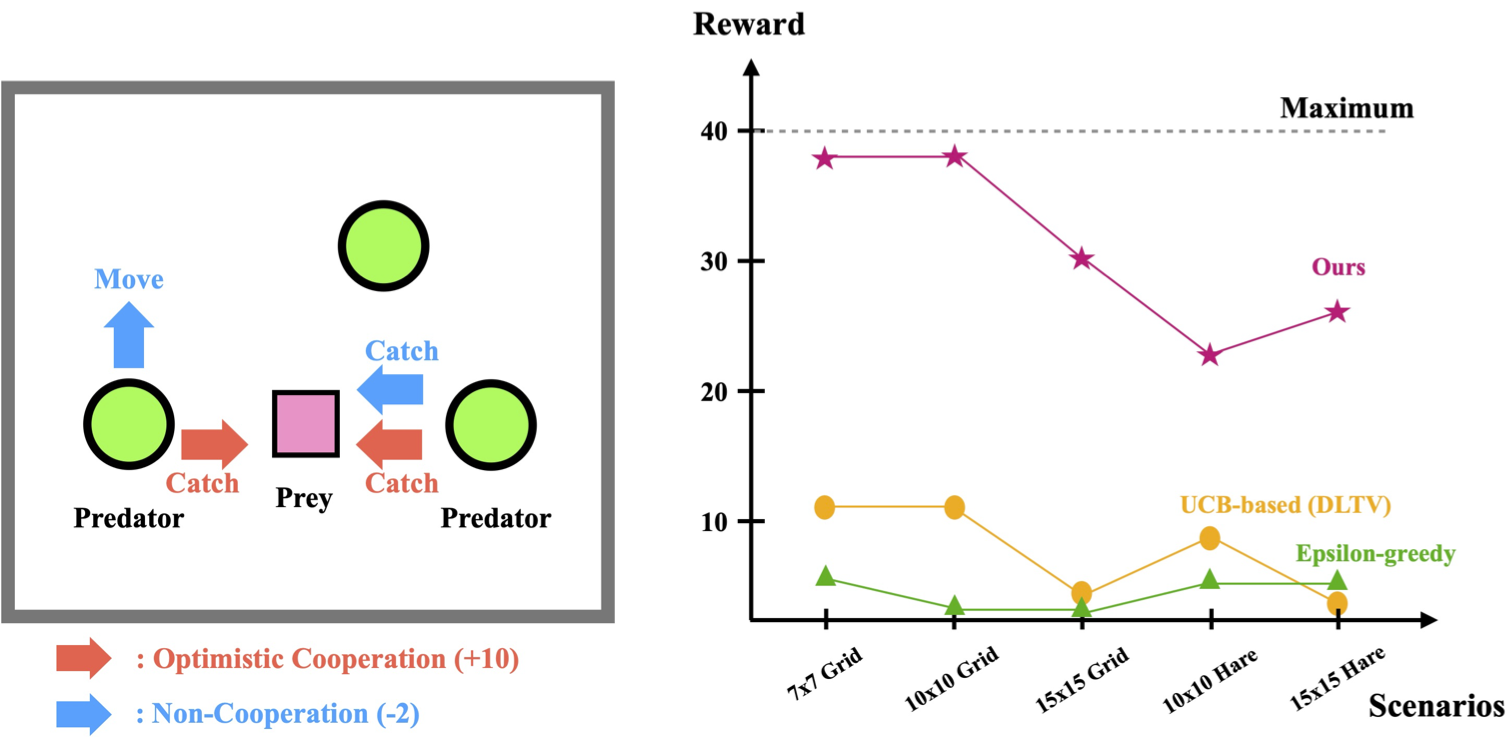}
    \hspace{-0.35cm} 
    \begin{subfigure}{0.45\linewidth}
      \centering
      \caption{} \label{part_a}
    \end{subfigure}
    \hspace{0.35cm} 
    \begin{subfigure}{0.45\linewidth}
      \centering
      \caption{} \label{part_b}
    \end{subfigure}
  \caption{Motivation, Predator \& Prey \cite{bohmer2020deep}. (a) represents how the environment works. (b) exhibits the reward per episode for each scenario. The lines are the mean of 6 random seeds.}
  \label{fig:toy_example}
\end{figure}

Reinforcement Learning (RL) \cite{sutton2018reinforcement} has been successfully used in various domains, such as robotics, autonomous driving, video games, economy, and operations research. Multi-agent reinforcement learning (MARL) \cite{sunehag2017value, rashid2018qmix, sun2021dfac, son2021disentangling}, which is an extension from the single-agent setting to the multi-agent setting, is in the spotlight because it can solve the complexity of a more realistic environment than single-agent learning. 
However, the behaviors of MARL algorithms are often very unpredictable because the actions chosen by each agent may influence other agents.
As the complexity of simulators evolves, the unpredictability makes it difficult for algorithms to approximate the exact state-action value. To address this environmental uncertainty, distributional variants of deep RL algorithms \cite{bellemare2017distributional, dabney2018distributional, dabney2018implicit} have been adopted in MARL, leading to the state-of-the-art performance in multi-agent settings such as the StarCraft Multi-Agent Challenges (SMAC) \cite{samvelyan2019starcraft}.
The distributional form of the state-action value reflects the aleatoric uncertainty arising from stochastic environments, multiple agents, and variances of reward distribution.
When representing state-action value as a distribution, there are two important features: \textit{variance} and \textit{risk}.
The \textit{variance} per action reflects the amount of uncertainty associated with parametric and intrinsic factors when an agent acts.
Thus, choosing actions with high variance is considered an optimistic approach that has the potential for a high return and is used for exploration \cite{auer2002using, mavrin2019distributional, nikolov2018information}.
The concept of \textit{risk} has its roots in economics and the stock market where prudent or audacious decisions are required.
It has been applied to RL in which the agent selects actions based on their risks; some approaches include Risk-Sensitive RL \cite{neuneier1998risk} and Safe RL \cite{garcia2015comprehensive}. 

In this study, we employ distributional RL to address one of the most fundamental challenges in RL, the exploration \& exploitation tradeoff.
Exploration collects informative samples, whereas exploitation exploits the (estimated) samples or actions. 
In the early stages, it is more advantageous to train agents with exploratory behavior, and in the later stages start to gradually converge it towards exploitation.
In multi-agent settings, the problem of exploration is more complicated due to the intrinsic uncertainty that arises from the multiple agents and unpredictable transition probability, which can be formalized using the Partially Observable MDP (POMDP) \cite{kaelbling1998planning}. 
Previous works on distributional MARL algorithms either rely on $\epsilon$-greedy \cite{bellemare2017distributional, dabney2018distributional, dabney2018implicit, yang2019fully, luo2021distributional} or UCB-based methods \cite{mavrin2019distributional, zhou2021non, cho2022distributional}, both of which are inappropriate since they do not take cooperative strategies into account.
However, distributional MARL algorithms still rely on the $\epsilon$-greedy exploration \cite{son2021disentangling, sun2021dfac, qiu2021rmix}, whereas numerous studies have proposed exploration strategies for distributional RL.

Figure \ref{fig:toy_example} shows the environment and performance results of Predator \& Prey \cite{bohmer2020deep} in the grid-world setting, which serves as an illustration of the importance of cooperative exploration.
Predators, which are agents, get a reward when they capture prey. When two predators catch a single prey at the same time, they receive a reward of +10 and a penalty of -2 when catching a prey solely. After two predators simultaneously capture a prey, the predators are immobile and eliminated. Due to the danger
of obtaining a negative reward, predators must locate and capture prey and work with other agents to maximize the reward. As shown in Figure \ref{fig:toy_example}(b), $\epsilon$-greedy and UCB-based explorations (DLTV)  \cite{mavrin2019distributional} shows low performance. The result indicates that exploration methods typically employed in distributional MARL are ineffective in multi-agent environments where cooperation between agents is necessary.
To overcome the unpredictability of the environment, learning to cooperate between agents requires cooperatively optimistic exploration.

In this paper, we present \textbf{R}isk-based \textbf{O}ptimistic \textbf{E}xploration (ROE), a method compatible with any existing distributional MARL algorithms, that leads to cooperatively optimistic behavior by shifting the sampling region of distribution.
In this context, \textit{distribution} is the output of any distributional RL algorithm, which is precisely the inverse CDF of the Q-value.
The domain and range of the inverse CDF are referred to as \textit{quantile fractions} and \textit{quantile}, respectively.
In the initial phase of training, for instance, we take expectations from the upper quantiles of state-action values, which lead to risky actions in pursuit of high reward, and gradually shift the sampling region of quantiles to the entire distribution.
By doing so, we ensure that each agent is exposed to the same overall level of risk, compelling them to take identically optimistic actions that induce cooperation.
As shown in Figure \ref{fig:toy_example}, our strategy, ROE, beats other considered exploration methods in which agents explore the optimal reward collaboratively.
In addition, we conduct studies on the standard MARL benchmark, SMAC \cite{samvelyan2019starcraft}, which is a cooperative setting that is much more complicated than the Predator \& Prey. 
Experiments are conducted using the state-of-the-art distributional MARL algorithms (DMIX, DRIMA) with our ROE as a plug-in.
The results demonstrate that our strategy outperforms other exploration methods by a large margin.
We summarize our contributions as follows:
\begin{itemize}
    \item We propose a novel risk-based exploration for cooperative multi-agent settings that can be used as a plug-in for any existing distributional MARL algorithms.
    \item We conduct a comprehensive evaluation of our method in MARL environments and demonstrate substantial performance improvement when cooperative exploration is required.
\end{itemize}


\section{Backgrounds}
\label{sec:backgrounds}

\subsection{Distributional Reinforcement Learning}
\label{sec:distributional_rl}

In reinforcement learning, the environment is often described by the Markov Decision Process (MDP), given by a tuple $\langle {X, A, P, R, \gamma} \rangle$. Here, $P(x'|x,a):X \times  A \times X\to[0, 1]$ is a transition probability function where $x'$ is the next state given a current state $x$ and action $a$. An agent in MDP receives rewards as the reward function $R(x, a) : X \times A \to \mathbb{R}$. $\gamma \in [0,1)$ is the reward's discount factor.
The learner's goal is to find an optimal policy $\pi$ maximizing the cumulative rewards $G_{\pi} = \sum^{\infty}_{t=0} \gamma^{t} R(x_{t}, a_{t})$ with a policy $a_{t} \sim \pi(\cdot|x_{t})$ that outputs an action distribution given a state.

Unlike traditional approaches to RL, distributional RL generates outputs as a distributional form of action.
Compared to a scalar-valued reward, a distributional form of the reward gives a much richer structure to the underlying environment. Note that the scalar-valued reward is the expectation of the reward distribution.
In this framework, the reward function becomes the reward distribution $R$, and the $Q$ function becomes a quantile function $Z$. We treat the expectation $\mathbb{E}[Z(x, a)]$ as the traditional $Q(x, a)$ value. The corresponding distributional Bellman equation is defined as follows~\citep{bellemare2017distributional} : 
\begin{gather}
\label{eqn:distributional_bellman_equation}
    \forall (x,a)\in X\times A : Z(x, a) \stackrel{d}{=} \mathcal{T}Z(x, a) := R(x, a) + \gamma Z(x',a')\,,\\ 
\text{where}\: x'\!\!\sim\! P(\cdot | x,a), a'\!\!\sim\! \pi(\cdot|x').\nonumber
\end{gather}
The mapping between distributions $\mathcal{T}$ is called the distributional Bellman operator \cite{bellman1966dynamic}. This distributional RL framework with the operator $\mathcal{T}$ is being widely studied, both theoretically \citep{bellemare2017distributional} and empirically \cite{dabney2018implicit}.

Categorical DQN \citep{bellemare2017distributional} gained popularity due to its superior performance in the Arcade Learning Environment (ALE) based on the Atari 2600 \citep{bellemare13arcade}. They output the return distribution given a state and an action by fixing the return values (known as atom or support) and approximating each return value's likelihood. The authors used the projected Kullback-Leibler (KL) divergence metric for loss functions using the shifted return values resulting from the added reward and $\gamma$. 
QR-DQN \cite{dabney2018distributional} fixes the distribution of the return as uniform and approximates return values with quantile regression.
They proved that the distributional Bellman operator is a $\gamma$-contraction w.r.t. the metric $\bar{d}_p$, which is the maximal form of the Wasserstein metric $W_p$:
 \begin{equation}
    \begin{aligned}
    \label{eqn:wasserstein}
     \bar{d}_p(Z_1, Z_2) = \sup_{\mathclap{{x\in X\,,a\in A}}}\quad\underbrace{\left ( \int_{0}^{1} \left| F_{Z_1(x,a)}^{-1}(\tau ) - F_{Z_2(x,a)}^{-1}(\tau )\right|^{p}d\tau  \right )^{1/p}}_{=:W_p(Z_{1}(x,a)\,,Z_{2}(x,a)) }
    \end{aligned}
\end{equation}
where the inverse CDF $F_{Y}^{-1}$ of a random variable \textit{Y} can be written as,
 \begin{equation}
    \begin{aligned}
    \label{eqn:inverseCDF}
     F_{Y}^{-1}(\tau ) := \textrm{inf}\left\{y \in \mathbb{R} : \tau \leq F_{Y}(y ) \right\}\,.     
    \end{aligned}
\end{equation}
Furthermore, they proposed the optimization method using the distributional TD-error $\delta_{\tau \tau '}$:
 \begin{equation}
    \begin{aligned}
    \label{eqn:distributional_td-error}
     \delta_{\tau \tau '} = R(x, a) + \gamma F_{Z_\theta(x', a')}^{-1}(\tau') - F_{Z_\theta(x, a)}^{-1}(\tau)\,,
    \end{aligned}
\end{equation}
where $x'\!\!\sim\! P(\cdot | x,a)$ and $a'\!\!\sim\! \pi(\cdot|x')$. In QR-DQN, each distribution of returns per action can be defined by a linear combination of Dirac measures as follows.
 \begin{equation}
    \begin{aligned}
    \label{eqn:dirac_delta}
     Z_{\theta }(x, a) := \frac{1}{N}\sum_{i=1}^{N}\delta_{\theta _i}{(x, a)}\,.
    \end{aligned}
\end{equation}
Here, $\theta_{i}$ and \textit{N} represents the return value and the number of return values each.
IQN \cite{dabney2018implicit} does not fix the probability of distribution and randomly selects the quantile fractions from a uniform distribution, $\mathcal{U}[0, 1]$. 
Distributional RL \cite{dabney2018distributional, dabney2018implicit} uses Huber \cite{huber1992robust} quantile regression loss $\rho^{k}_{\tau}$ defined as:
 \begin{equation}
    \begin{aligned}
    \label{eqn:huber_loss}
    & \rho^{k}_{\tau}(\delta_{\tau \tau '}) = |\tau - \mathbb{I}\{\delta_{\tau \tau '}<0\}|\cdot\mathcal{L}_{k}(\delta_{\tau \tau'}),
    \end{aligned}
\end{equation}
where
 \begin{equation}
\label{eqn:huber_loss_2}
\mathcal{L}_{k}(\delta) =
\begin{cases}
    \frac{1}{2k}\delta^{2} & \mathrm{if} \,\, |\delta|\leq k\\
    |\delta|-\frac{1}{2}k & \mathrm{otherwise}
\end{cases}
\end{equation}

Based on the distributional Bellman operator,  NDQFN \cite{zhou2021non} and SPL-DQN \cite{luo2021distributional} utilized a monotonic structure design to guarantee non-decreasing return values according to the arising quantile fractions.
Instead of sampling quantile fractions from a distribution, FQF \cite{yang2019fully} samples quantile fractions as a parameterized model. Recently, risk-sensitive RL has been conducted based on the distributional RL due to its ability to handle quantile fractions \cite{lim2022distributional}.

\subsection{Risk-Sensitive Policy}
Generally, in risk-sensitive RL \cite{neuneier1998risk}, risk levels can be divided into three sections: risk-averse, risk-neutral, and risk-seeking.
Due to the variation in action space, a risk-sensitive policy can be read differently based on the context. Nonetheless, in this section, we will describe the general concept of risk-related policy.
A risk-averse policy can be interpreted as acting with the highest state-action value among the worst-case scenarios per action. A risk-seeking policy entails selecting the same action as a risk-averse policy but based on the best-case scenario. Risk-neutral policy positions amid risk-averse and seeking policy positions.


\subsection{Multi-Agent Reinforcement Learning}
We now review some recent developments in deep MARL.
VDN \cite{sunehag2017value} considers a joint state-action value ($Q_{joint}$), which is just the summation of all agents' state-action values ($Q_{agent}$).
QMIX \cite{rashid2018qmix}, which is the most well-known algorithm in MARL, maintains monotonicity in incorporating $Q_{agent}$ to $Q_{joint}$.
Recently, some MARL algorithms adopted a distribution-based architecture. DMIX \cite{sun2021dfac} integrated distributional RL and MARL via mean-shape decomposition, which is inspired by QMIX \cite{rashid2018qmix}.
In DMIX, a small number of quantile fractions are sampled from $\mathcal{U}[0, 1]$, resulting in a distorted uniform distribution rather than a perfect one. However, when considering the entire episode, the sampling quantile fractions approach the uniform distribution $\mathcal{U}[0, 1]$.
DRIMA \cite{son2021disentangling} divided risk sources into cooperative and environmental risks, and injected risk levels into the agent utility function and centralized utility function differently according to the environment.
They employed the architecture of the QTRAN \cite{son2019qtran} for the overall structure and the QMIX for the hypernetwork.
RMIX \cite{qiu2021rmix} adopted the Conditional Value at Risk (CVaR) as a surrogate of joint state-action value $Q_{joint}$ and developed a model that adaptively estimates CVaR at every step.

\subsection{Exploration in RL}

Exploration is the key problem in reinforcement learning. It has an inherent trade-off with exploitation, which is significant as it impacts the sample efficiency of RL, and can be affected by many factors such as sparse or delayed reward, large state \& action space, and more.
Various algorithms, such as the $\epsilon$-greedy \cite{sutton2018reinforcement}, Boltzmann exploration \cite{sutton90integratedarchitectures}, noise perturbation \cite{fortunato2017noisy}, and intrinsic motivation \cite{bellemare2016unifying, pathak2017curiosity, burda2018exploration}, have been developed to solve this problem. After being developed using deep learning-based distributional RL \cite{bellemare2017distributional}, several distributional RL exploration methods have utilized the distributional property.
DLTV \cite{mavrin2019distributional} uses the QR-DQN algorithm \cite{dabney2018distributional} with optimistic action selection via the return distribution's left truncated variance.
Analogous to the Upper Confidence Bound (UCB) approach in bandits literature
, QR-DQN suppresses the intrinsic uncertainty by decaying the bonus such that only parametric uncertainty is utilized.
The Distributional Predict Error (DPE) algorithm \cite{zhou2021non} utilizes Random Network Distillation \cite{burda2018exploration} to generate two identical architectures with randomly initialized parameters and use their Wasserstein distance to measure an action's novelty given a state.

Although these methods are effective in the single-agent domain, na\"{i}vely applying them separately and independently to each agent in the MARL setting is bound to result in suboptimal performance.
This is because the agents mutually influence one another, creating additional uncertainty that has to be taken into account.
One way of accounting for such uncertainty is to require cooperative behavior between agents.
Recently, there have been some works on ensuring cooperation between the agents.
MAVEN \cite{mahajan2019maven} uses a hierarchical architecture to generate a shared latent vector for each agent to explore the space cooperatively.
CMAE \cite{liu2021cooperative} creates an exploration compartment for each agent that is not shared with other agents, drastically reducing the searching space via cooperative behavior.
However, none of these algorithms account for inherent uncertainty and are applied to other MARL algorithms.
\section{Risk-based Optimistic Exploration}
\label{sec:method}
We propose a model-agnostic risk-based optimistic exploration method for a cooperative multi-agent setting by shifting the sampling region of the state-action value's distribution. 
In Section \ref{sec3.1}, we first show the limitations of existing model-agnostic exploration methodologies and the conventional definition of risk for MARL.
In Section \ref{sec3.2}, we discuss how our methodology works to achieve cooperative optimism by satisfying the $\gamma$-contraction in the distributional Bellman operator.

\subsection{Risk in MARL}
\label{sec3.1}
\begin{figure}[!ht]
  \centering
  \includegraphics[width=\linewidth]{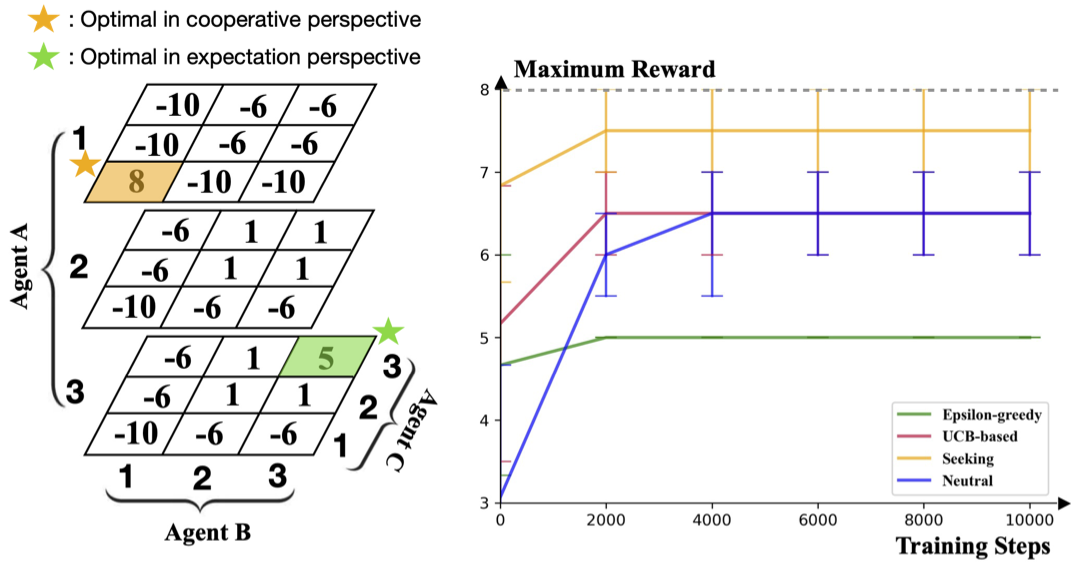}
  \hspace{-1.5cm} 
    \begin{subfigure}{0.45\linewidth}
      \centering
      \caption{} \label{part_a_2}
    \end{subfigure}
    \hspace{0.35cm} 
    \begin{subfigure}{0.45\linewidth}
      \centering
      \caption{} \label{part_b_2}
    \end{subfigure}
  \caption{Toy example on 1-Step Payoff game. (a) represents the matrix game, and (b) shows the reward of the each approach.}
  \label{fig:matrix}
  \Description{}
\end{figure}

Figure \ref{fig:matrix} shows a clear performance gap between different exploration methods for our considered toy example.
In this environment, the true state-action values for all agents are $Q(x, a_{1})=-56$, $Q(x, a_{2})=-30$, and $Q(x, a_{3})=-26$ respectively.
Although the maximum reward $8$ is given when all agents choose $a_{1}$,
for each agent, $a_{3}$ is the best action to maximize its individual reward in expectation.
Therefore, unless the agents are altruistic, each agent will choose $a_3$, which is its individually optimistic action.
In order to obtain the largest reward $8$, the environment must force the agents to pick the cooperatively optimistic action, $a_1$.
We use the DMIX algorithm in this environment and sample $\tau$ from $\mathcal{U}[0, 1]$, which is the default risk-neutral setting.
As shown in Figure \ref{fig:matrix}(b), $\epsilon$-greedy exploration fails to identify the maximum reward until the end of the training and instead, obtains a suboptimal reward $5$ in expectation.
Moreover, it can be seen that $\epsilon$-greedy exploration even performs worse than the risk-neutral setting, which suggests that the $\epsilon$-greedy approach hinders cooperative exploration.
The multi-agent version of DLTV (UCB-based method) receives a reward of $8$ half of the time and a reward of $5$ the other half of the time.
This is because DLTV compels agents to choose optimistic, non-cooperative actions, which results in optimal value when they are fortunate. We also consider a simple risk-based method (risk-seeking) for optimistic action.

After the quantile fractions' sampling region is specified, the risk-based distributional RL selects an action as follows.
\begin{eqnarray}
    \begin{aligned}
    \label{eqn:scheduling}
a^* = \argmax_{a\in A} \mathop{\text{\Large{$\mathbb{E}$}}}_{\tau\sim \mathcal{U}[\alpha,\beta]}\left[F^{-1}_{Z(x, a)}(\tau)\right]
    \end{aligned}
\end{eqnarray}

Usually, $\alpha$ and $\beta$ are set to $0$ and $1$ each to utilize full distribution, but this isn't always the case.
For instance, assuming that we want $0.5 < \alpha < \beta$ and that the distribution is symmetric about $0.5$, we get a general inequality as follows:
\begin{eqnarray}
    \begin{aligned}
    \label{eqn:overestimate}
    \mathop{\text{\Large{$\mathbb{E}$}}}\left[Z(x, a)\right] \leq \mathop{\text{\Large{$\mathbb{E}$}}}_{\tau\sim \mathcal{U}[\alpha, \beta]}\left[F^{-1}_{Z(x, a)}(\tau)\right]
    \end{aligned}
\end{eqnarray}
If the given $F^{-1}$, state and action are identical, the agent overestimates the state-action value as shown in Equation \eqref{eqn:overestimate} with upper quantile fractions.
Therefore, the agents now choose risk-seeking (optimism) action, leading to superior performance than risk-neutral policy when cooperative behavior is desired as shown in Figure \ref{fig:matrix}.
Here, we set $\alpha$ and $\beta$ to 0.75 and 1.0 to implement a risk-seeking policy.
Indeed, as shown in Figure \ref{fig:matrix}(b), the risk-seeking approach significantly outperforms $\epsilon$-greedy and DLTV, reaching the maximum reward more often and having a greater reward in expectation as well. 

Although the previous toy example suggests that risk-seeking always yields superior performance via the effect of cooperative optimism,
this isn't generally the case in a more complex and long-episodic environment.
In such environments, in contrast to the 1-step payoff game, continually seeking a high reward is not exploitation.
As the long-term episode requires the agents to decide their actions consecutively, the only-seeking method's cooperative strategy is broken. The agents have to exploit the estimated samples from the optimistic actions rather than explore only seeking behavior. 

\subsection{Cooperative Optimism with Risk Scheduling}
\label{sec3.2}

We propose ROE, \textbf{R}isk-based \textbf{O}ptimistic \textbf{E}xploration, which addresses the difficulties of multi-agent environments requiring cooperation as discussed in Section \ref{sec3.1}.
We achieve cooperative optimism in a multi-agent setting by endowing each agent with an identical risk level, hence inducing similar behaviors across the agents.
By imposing a high-risk level at the initial phase, (e.g., $\tau \sim \mathcal{U}[0.75, 1]$), we make the agents choose informative action in a cooperative manner.
We then gradually update the sample region, starting from the upper domain $\tau \sim \mathcal{U}[0.75, 1]$ to the full domain $\tau \sim \mathcal{U}[0, 1]$.
Lastly, the agents exploit the estimated samples of the entire distribution.

We allow agents to explore cooperatively optimistic actions, gradually exploiting the optimistically estimated samples using Equation \eqref{eqn:scheduling} where $\alpha$ and $\beta$ adjust the risk levels (confidence bound of distribution) and keep changing through the scheduling steps as illustrated in Algorithm \ref{sudo_code}.

\begin{algorithm}[tb]
\caption{ROE [Linear scheduling]}
\label{sudo_code}
\begin{algorithmic}[1] 
\REQUIRE 
\STATE $k \leftarrow$ scheduling time steps
\STATE $\omega_{0} \leftarrow$ initial risk level, $\omega_{k} \leftarrow$ final risk level

\textcolor{gray}{\# We set risk level interval to [-1, 1] for the convenience of computation. Risk level 1 (extreme seeking), 0.5, 0 (neutral), -0.5, -1 (extreme averse) means sampling quantile fractions from $\mathcal{U}[1, 1], \mathcal{U}[0.5, 1], \mathcal{U}[0, 1], \mathcal{U}[0, 0.5], \mathcal{U}[0, 0]$ each. Therefore, if we set $\omega_{0} = 1$ and $\omega_{k} = 0$, then it means that I will schedule the risk level from risk-seeking to neutral.}
\ENSURE 
\STATE $\omega_{t} \leftarrow$ current risk level $(= [\alpha_{t}, \beta_{t}]), \,\, 0 \leq \alpha_{t} \leq \beta_{t} \leq 1$
\STATE Risk-scheduling size $\delta$ is  $\delta = \frac{\omega_{0} - \omega_{k}}{k}$
\STATE Store random trainsition $(x_{t}, a_{t}, r_{t}, x_{t+1})$ in ReplayBuffer $\mathcal{D}$ for short learning time.
\STATE $\omega_{t} \leftarrow \omega_{0}$
\WHILE{$t < T$}
\STATE Select an action, $a_{t}=\mathrm{argmax}_{a\in A} \mathbb{E}_{\tau\sim \mathcal{U}[\alpha_{t},\beta_{t}]}[F^{-1}_{Z(x, a)}(\tau)]$
\STATE Execute an action $a_{t}$ and observe $r_{t}$ and $x_{t+1}$
\STATE Store transition $(x_{t}, a_{t}, r_{t}, x_{t+1})$ in ReplayBuffer $\mathcal{D}$
\STATE Sample transition batch from ReplayBuffer $\mathcal{D}$
\STATE $\mathcal{L}_{k} = \frac{1}{N} \sum\limits_{i=1}^N\sum\limits_{j=1}^N \rho^{k}_{\tau_{i}}(\delta_{\tau_{i} \tau_{j}} '^{t}), \,\,\, \tau_{i} , \tau_{j} ' \sim \mathcal{U}[\alpha_{t}, \beta_{t}]$
\IF{$t \leq k$}
\STATE $\omega_{t+1} \leftarrow \omega_{t} - \delta$
\ELSE
\STATE $\omega_{t+1} \leftarrow \omega_{k}$
\ENDIF
\ENDWHILE
\end{algorithmic}
\end{algorithm}

\subsubsection{Dynamics of risk-scheduling}

\newcommand{\fixedp}{{Z^{*}\!\!\!}}
ROE shifts risk levels from the \textit{seeking} to specific levels.
Like to previous works~\cite{bellemare2017distributional,Keramati2020BeingOT}, our method is equivalent to iterating a finite sequence of operators $\{\mathcal{T}\circ \Pi_{\alpha_t, \beta_t} \}_{t=1}^T$, where $\Pi_{\alpha_t, \beta_t}$ is the uniform projection on the quantile range $[\alpha_t,\beta_t]$.
We discuss the contraction property of the  distributional Q function and apply the distributional optimality operator $\mathcal{T}=\mathcal{T}^\pi$ for a greedy policy $\pi$~\cite{bellemare2017distributional}.

We first note the non-expansive property of the projection operator in the following Lemma~\ref{lem:nonexpansive}.
\begin{lemma}[Non-expansiveness]
\label{lem:nonexpansive}
Let $\Pi_{\alpha, \beta}$ ($0\leq \alpha < \beta \leq 1$) be the transformation on the random variable, defined by the quantile function or inverse CDF as
\begin{align*}
F^{-1}_{\underline{\Pi_{\alpha,\beta}Z(x,a)}}(&\tau) =  F^{-1}_{Z(x,a)}\left((\beta-\alpha)\tau + \alpha \right)\,,\\
\mathrm{where}\:\: &\tau\in[0,1],\,(x,a)\in X\times A\,.
\end{align*}
Then the $\Pi_{\alpha,\beta}$ is non-expansive on the metric $\bar{d}_\infty$:
\[
\bar{d}_\infty(Z_1,Z_2) = \sup_{\mathclap{{x\in X\,,a\in A}}}\quad\, \esssup_{\tau\in[0,1]} \left|F^{-1}_{Z_1(x,a)} (\tau) - F^{-1}_{Z_2(x,a)}(\tau)\right|\,.
\]
\end{lemma}
Therefore, if the distributional Bellman operator with greedy policy $\mathcal{T}$ is a $\gamma$-contraction, so is $\mathcal{T}\!\circ\!\Pi_{\alpha,\beta}$ on $\bar{d}_\infty$, for fixed $\alpha$ and $\beta$. Furthermore, by the Banach fixed point theorem, there also exists a unique fixed point $Z_{\alpha,\beta}$ for $\mathcal{T}\circ\Pi_{\alpha,\beta}$.
Each fixed point, which is precisely the distributional $Q$ function, reflects the various risk level by allowing agents to behave differently.

From these observations, we propose a scheduling method to allow the agents to various risk-sensitivity.
Since the iterating operator changes with time, the procedure is governed by the temporal evolution of the operator.
This is especially true when $\alpha_t$ and $\beta_t$ change at a rate of $o(T^{-1})$. 
In such cases, a mere convergence result does not provide much information. Instead, we show that the distance between the $t$-th step and the fixed point $\fixedp_{\alpha_t,\beta_t}$ can be bounded as shown in the following proposition :

\begin{proposition}
Let us consider the iterative process $Z_t$ $\leftarrow$ $\mathcal{T}\circ\Pi_{\alpha_t, \beta_t}(Z_{t-1})$, and denote $\fixedp_{\alpha_t,\beta_t}$ as the unique fixed point of $\mathcal{T}\circ\Pi_{\alpha_t, \beta_t}$. Then we have the upper bound between the distance of the $t$-th state and the fixed point of $t$-th operator as:
\begin{align*}
\bar{d}_\infty\big(Z_t,\,\fixedp_{\alpha_t,\beta_t}\big) \leq &\sum_{i=1}^{t-1} \gamma^{t-i} \bar{d}_\infty\big(\fixedp
_{\alpha_{i},\beta_{i}},\,\fixedp_{\alpha_{i+1},\beta_{i+1}}\big) + \gamma^t \bar{d}_\infty\big( Z_0,\,\fixedp_{\alpha_{1},\beta_{1}}\big) \,.
\end{align*}
\end{proposition}
The upper bound is a weighted combination of the $\bar{d}_\infty$-distance between the neighboring fixed points.
Intuitively, recent information has a more significant influence, which exponentially decreases with its age by a factor of $\gamma$, the discount factor.
One important observation is that if $(\alpha_t,\beta_t)$ changes moderately towards $(\alpha,\beta)$, $Z_t$ remains close to ${Z^{*}\!\!\!}_{\alpha,\beta}$ because the distance between the fixed points will be close.

\section{Experiments}

\begin{figure}[!t]
  \centering
  \includegraphics[width=\linewidth]{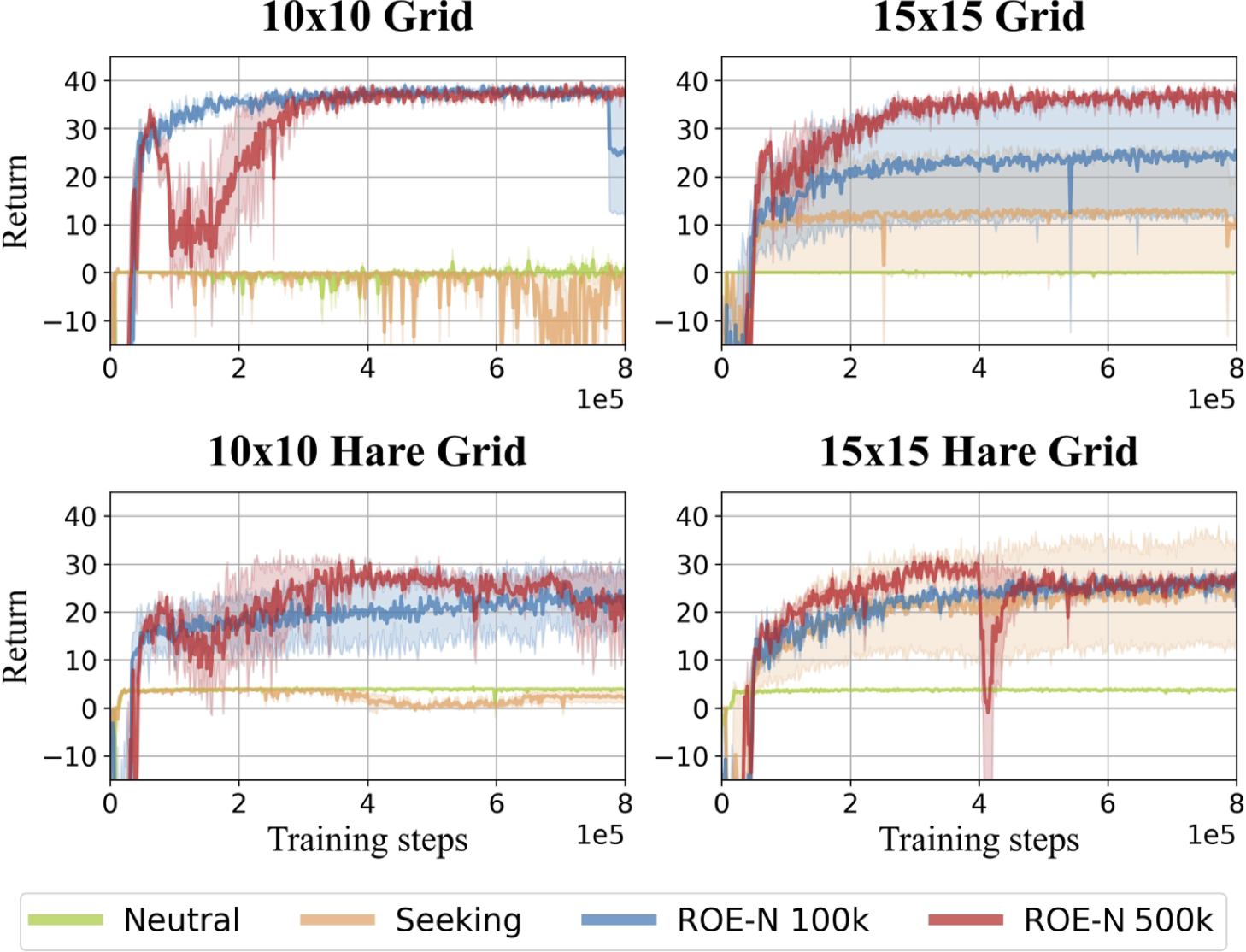}
  \caption{Episode return of DMIX in Predator \& Prey experiments. The lines are the mean of 3 random seeds with shaded areas representing a confidence interval of 25\% to 75\%. The numbers represent risk-scheduling steps.}
  \label{fig:predator_results}
  \Description{}
\end{figure}
\label{sec:experiments}
For the experiments, we consider two variants of ROE: ROE-N refers to ROE that adjusts the risk level from risk-seeking to risk-neutral, and ROE-A refers to ROE that adjusts from risk-seeking to risk-averse.
We evaluate ROE-N and ROE-A in two cooperative multi-agent settings with high aleatoric uncertainties.
One is a Predator \& Prey environment and the other is the Starcraft Multi-Agent Challenges (SMAC).
As an ablation study, we also consider a single-agent setting that does not require a cooperative strategy.
Additional experimental details and results are provided in the Appendix \ref{sec:ablation}.

\begin{figure*}[!t]
  \centering
  \includegraphics[width=0.75\linewidth]{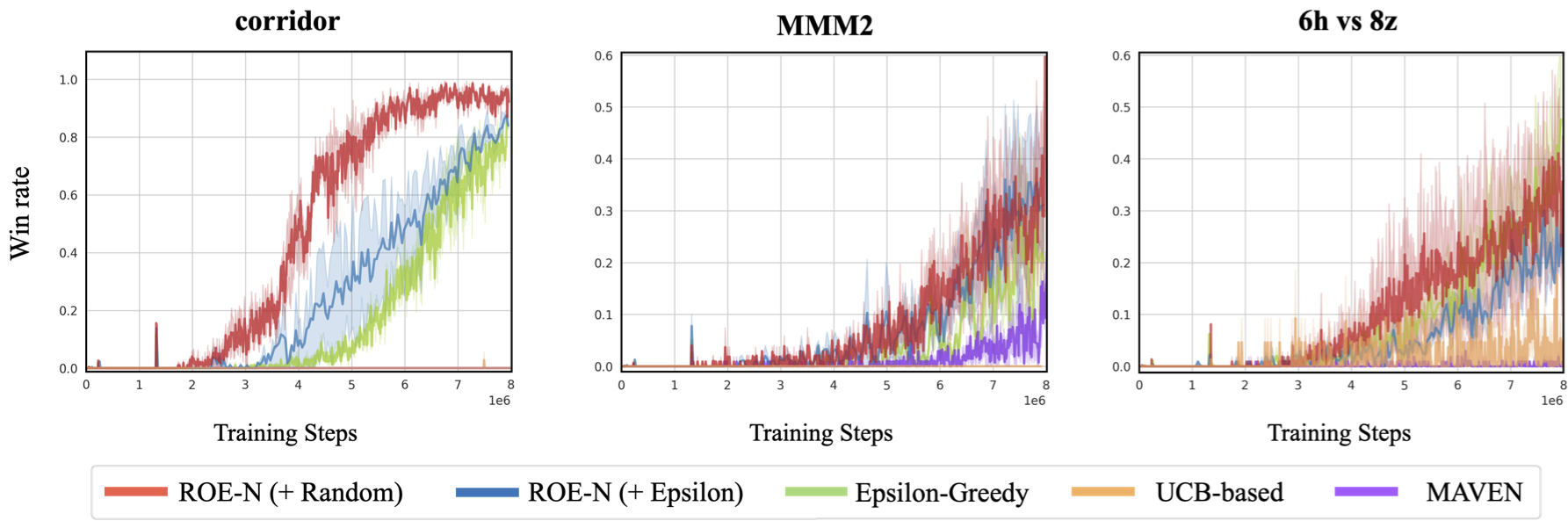}
  \caption{Comparison of exploration algorithms in Superhard scenarios. The lines are the mean of 3 random seeds using five parallel training with shaded areas representing a confidence interval of 25\% to 75\%. Baseline architecture is DRIMA.}
  \label{fig:exploration_results}
  \Description{}
\end{figure*}

\begin{figure*}[!t]
  \centering
  \includegraphics[width=0.75\linewidth]{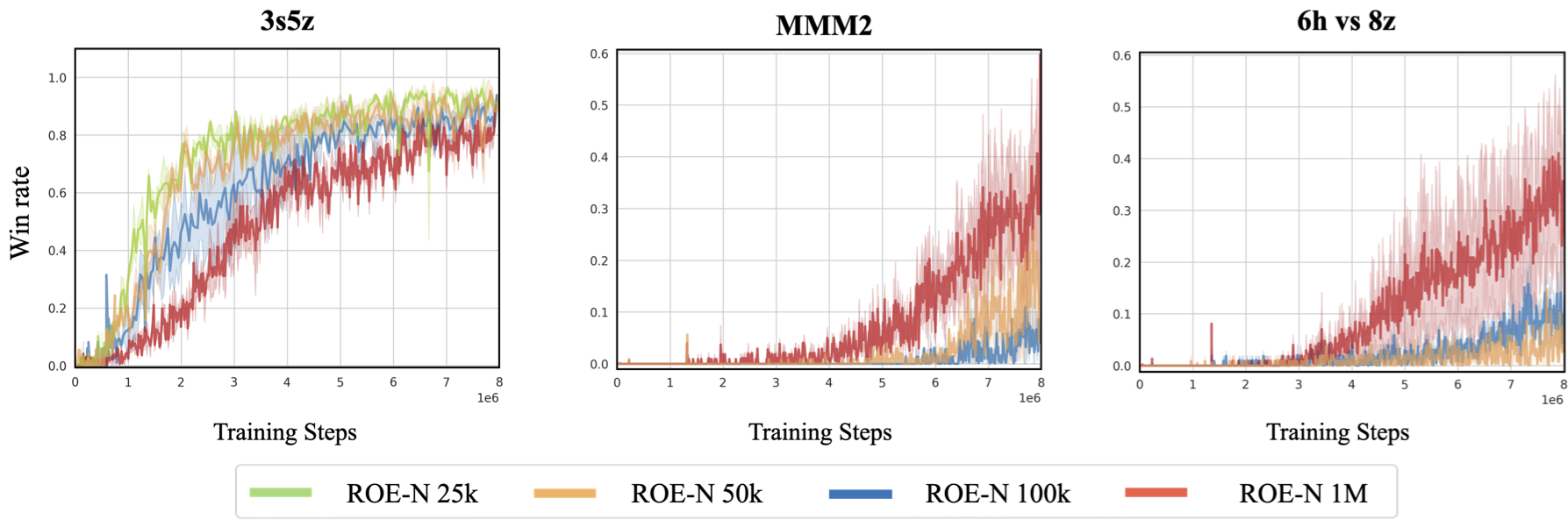}
  \caption{Performance sensitiveness of our method (ROE) according to the risk-scheduling steps on \textit{Easy} scenario (3s5z) and \textit{Superhard} scenarios (MMM2 \& 6h vs 8z). The lines are the mean of 3 random seeds using five parallel training with shaded areas representing a confidence interval of 25\% to 75\%. Baseline architecture is DRIMA.}
  \label{fig:sensitiveness}
  \Description{}
\end{figure*}

\subsection{Environments}
\textbf{Predator \& Prey} 
As dealt with previously in the Introduction section, Predator \& Prey \cite{bohmer2020deep} is a grid environment in which the 8 predators (agents) must capture 8 prey cooperatively.
The environment has inherent stochasticity as follows: with probability $0.1$ an "up" action will not be executed, and the transition of each predator is governed by a transition probability kernel $P(x'|x,a)$.
Each prey begins each episode at an arbitrary point and behaves in a random manner, resulting in an inability to remember sequences for agents.
Moreover, agents can observe only within two grids from them, which makes this environment a POMDP.
The environment thus requires cooperative strategies and optimistic exploration to capture the prey.
Additionally, in the harder scenario \texttt{Hare Grid}, there exist rabbits that are similar to prey in the way of giving rewards but provide reward 1.
Such rabbits are used as deceptive reward signals which hinder predators from capturing prey.

\textbf{StarCraft Multi-Agent Challenges (SMAC)}
For more complex POMDP multi-agent settings, we conduct experiments on SMAC environments \cite{samvelyan2019starcraft}, the standard cooperative multi-agent RL benchmark, with a focus on micro-management challenges.
Each SMAC environment consists of allies and enemies, each evaluating their win rate.
Allies are controlled by the MARL algorithms, while enemies are controlled by the original StarCraftII agents with a difficulty level 7 out of 10.
Allies receive the episode's reward of $200$ when they win a battle, as well as small rewards of 10 for killing an enemy and a payout equal to the amount of damage they dealt to adversaries.
To win a battle, agents must cooperate among themselves to manage their group behavior, like \textit{focusing fire} while not overkilling the enemies, or \textit{kiting} to lure the enemies and kill them one by one.
We report the results for \textit{SuperHard} scenario, where the importance of cooperation is crucial in winning.
The results for \textit{Easy} and \textit{Hard} scenarios are reported in Appendix \ref{sec:ablation}.

\textbf{Atari}
We evaluated the validity of our method in an Atari game \cite{bellemare13arcade,machado2018revisiting}, a single-agent setting where intrinsic uncertainty is very low and cooperation is not necessary.
Here, the environment is fully deterministic, and the reward consists of \{-1, 0, 1\}.
Specifically, the experiments are conducted in situations where complex exploration is needed \cite{taiga2021bonus}.

\begin{figure*}[!t]
  \centering
  \includegraphics[width=\linewidth]{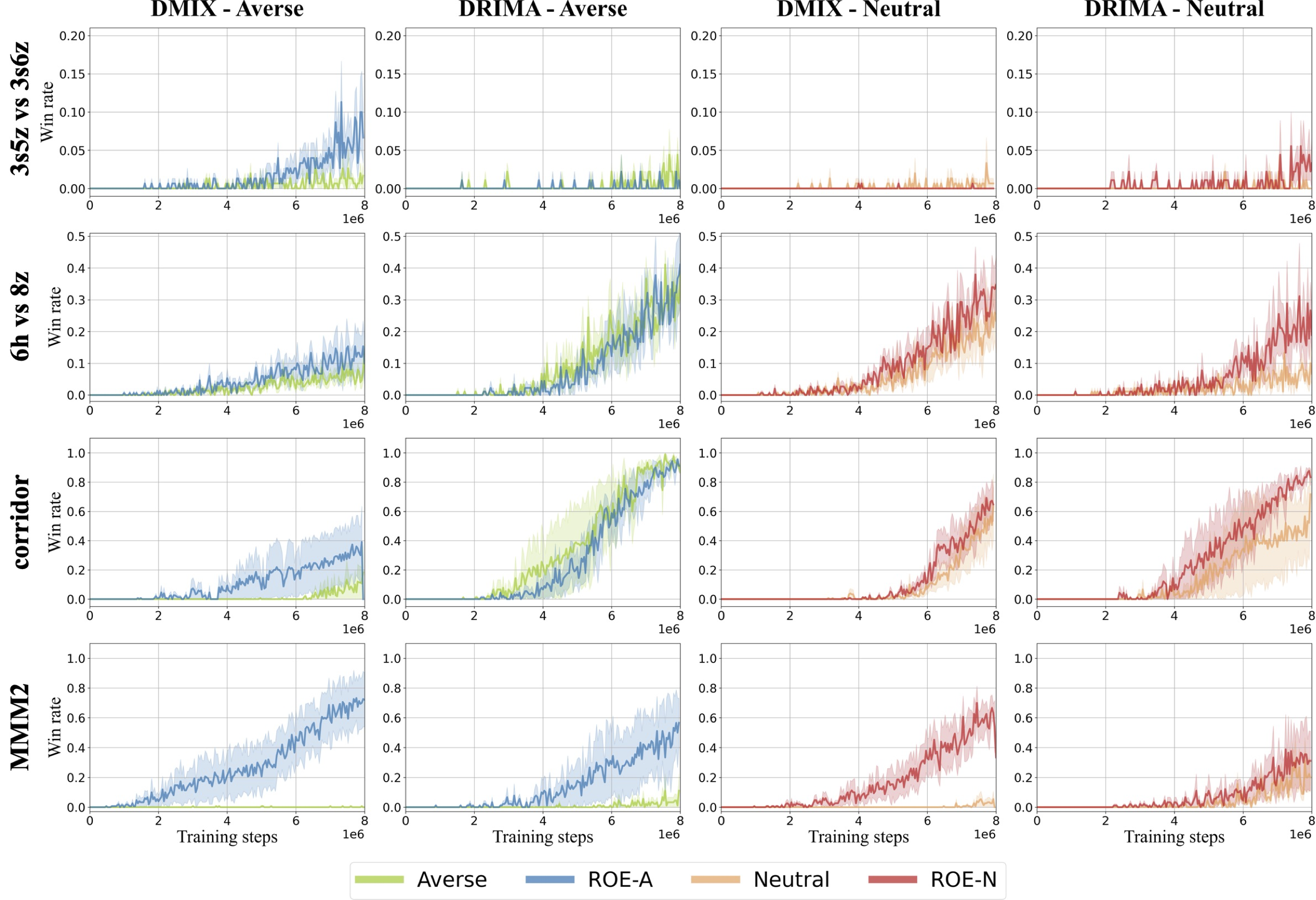}
  \caption{Win-rate Results of DMIX and DRIMA in Superhard scenarios. The label of X-axis and Y-axis represent algorithm - risk level and scenarios respectively. The lines are the mean of 5 random seeds in DMIX, 3 random seeds in DRIMA using five parallel training with shaded areas representing a confidence interval of 25\% to 75\%.}
  \label{fig:main_results}
  \Description{}
\end{figure*}
\subsection{Implementation}
For risk-based optimistic exploration, we shift the sampling region of distribution with \textit{linear} scheduling.
We plugged in our ROE method to IQN, DMIX, and DRIMA.
For IQN and DMIX, we define risk-averse, risk-neutral, and risk-seeking to be the sampling quantile fractions from $\mathcal{U}[0, 0.25]$, $\mathcal{U}[0, 1]$, and $\mathcal{U}[0.75, 1]$, respectively; for DRIMA, they were set to be the sampling quantile fractions from $\mathcal{U}[0, 0.1]$, $\mathcal{U}[0.4, 0.5]$, and $\mathcal{U}[0.9, 1.0]$.
To schedule risk in IQN and DMIX from seeking to neutral, we initialized $(\alpha, \beta) = (0.99, 1.0)$ in Equation \ref{eqn:scheduling} at first, for more optimistic exploration, which is generally set to (0.75, 1) for risk-seeking policy.
Then, quantile fractions 0.99 ($\alpha$) is linearly decayed to $0$.
When $\alpha$ becomes $0$, the risk level is positioned at risk-neutral, which samples quantile fractions from $\mathcal{U}[0, 1]$.
In DRIMA, we linearly shift the quantile sampling index from $\mathcal{U}[0.9,1.0]$ to $\mathcal{U}[0.4, 0.5]$ by an increment of $0.1$ to correspond to the architecture of the underlying algorithm.
Additional details are presented in Appendix \ref{sec:ablation}.
In SMAC environment, we evaluate both ROE-N and ROE-A.
This is because we have observed that, depending on the risk level of the agents, SMAC displays distinct behaviors that directly influence the win rate.

\subsection{Results}
In MARL experiments with high uncertainty levels where cooperation is necessary, our risk-based exploration yields a significant performance advantage over $\epsilon$-greedy, UCB-based exploration, and static risk level-based approaches.

\textbf{Predator \& Prey} 
We compare our method with static risk level-based approaches.
Figure \ref{fig:predator_results} shows the training curves of each algorithm.
Predators plugged with our method effectively resolve the problem that static predators could not.
In environments \texttt{10x10 Grid} and \texttt{15x15 Grid}, which lack deceptive reward compared to \texttt{Hare Grid}, risk-neutral predators could evade the negative reward that results from solely capturing, but they do not learn how to get a greater reward. Risk-seeking predators demonstrate moderately superior or even worse than risk-neutral predators. However, our method initially receives negative rewards but has cooperative optimism, which will be decayed to find better rewards, so they learn the appropriate methods and employ them effectively. In the setting that has a deceptive reward, \texttt{Hare}, shows similar results. Risk-neutral predators only capture rabbits (deceptive rewards), hence never capturing prey. It is easy for static risk-neutral predators to learn how to take rabbits but difficult to learn how to capture prey. Although risk-seeking predators perform similarly to ROE predators in the \texttt{15x15 Hare Grid}, they do poorly in the \texttt{10x10 Hare Grid}. However, predators appear to perform well with our ROE that maximizes rewards in all \texttt{Hare Grid} while effectively avoiding rabbits. This phenomenon illustrates that our methodologies are robustly operational, even in smaller or more challenging maps (i.e., \texttt{10x10 Hare Grid}) where it is easier to receive deceptive rewards.

In addition, for a more detailed explanation for comparing exploration methods, we use the experiment results in Introduction section. As shown in Figure \ref{fig:toy_example}(b), our method outperforms the other exploration methods with a significant performance gap. The failure of $\epsilon$-greedy exploration in this environment is due to the random exploration's discontinuity of preferable actions and using only expectation value to choose an action. UCB-based exploration demonstrates better exploration (sometimes reach to maximum reward when fortunate) than the $\epsilon$-greedy method, but it exhibits most of the failure in getting rewards. This is because UCB-based exploration, choosing based on the variance of reward distribution,  yields optimistic but non-cooperative action.


\begin{figure}[!t]
  \centering
  \includegraphics[width=\linewidth]{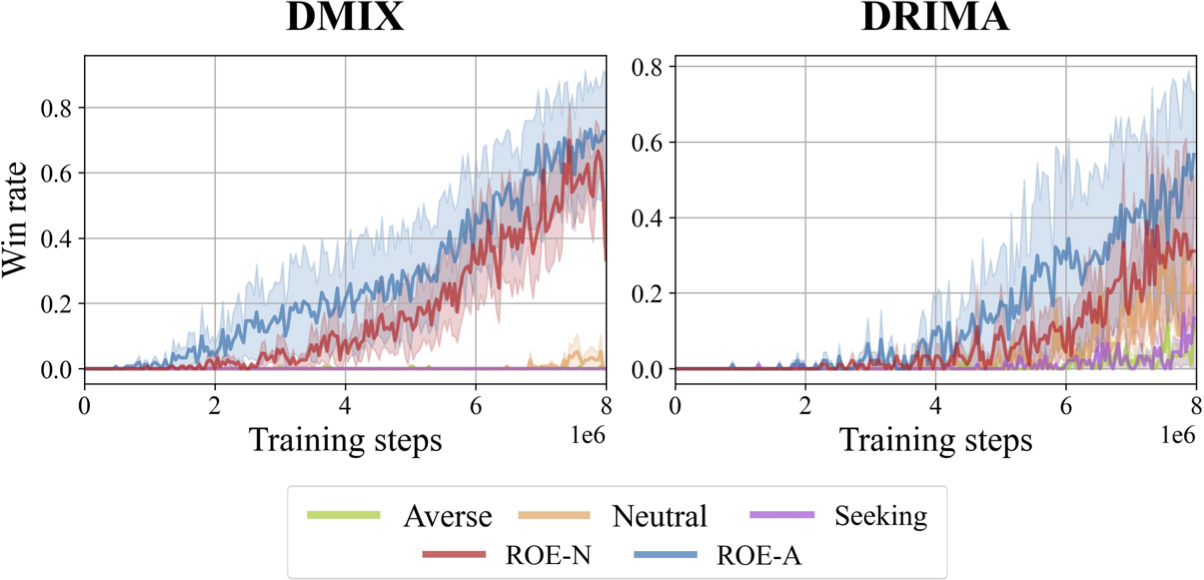}
  \caption{Win-rate Results of \texttt{MMM2} scenario in SMAC. The lines are the mean of 3 (left) and 5 (right) random seeds with shaded areas representing a confidence interval of 25\% to 75\%.}
  \label{fig:mmm2_results}
\end{figure}
\textbf{StarCraft Multi-Agent Challenges}
Our method results in considerable performance gains in SMAC. We compare the exploration methods ($\epsilon$-greedy, UCB-based, MAVEN\cite{mahajan2019maven}), which are applicable to any algorithms (except MAVEN) in \textit{Super Hard} scenarios where hard exploration is required. Additionally, to make sure monotonicity in the inverse CDF function (quantile function), we collect random samples in the very initial training phase for 50k steps by random or $\epsilon$-greedy action selector. For long-horizontal exploration, we searched exploration step in \{50k, 100k, 1M\} for our method and $\epsilon$-greedy exploration and showed the best performance among them. The hyperparameters used in UCB-based and MAVEN is that showing the best performance in their papers. As shown in Figure \ref{fig:exploration_results}, ROE with random (purple line) shows the best performance. Also, we compare ROE with a static risk-neutral, averse policies with comprehensive experiments. As depicted in Figure \ref{fig:main_results}, our learning curve converges faster and obtains a higher win rate in the majority of scenarios compared to static risk policies. 

The reason of the performance in Figure \ref{fig:exploration_results},\ref{fig:main_results} is that ROE collects the merits of risk-seeking and other risk-based policies by scheduling the risk levels in SMAC. Risk-seeking policies at an early stage enable allies to explore and identify cooperative winning strategies, such as running away for a moment or moving to weakened enemies to focus fire, more quickly by encouraging them to take cooperatively optimistic actions. In contrast, risk-averse policies generally encourage allies to focus mainly on attack, which is the best course of action in the worst-case scenario. Since decaying the risk level controls this trade-off effectively, ROE could achieve the best performance among our experiments.

As demonstrated in Figure \ref{fig:exploration_results}, similar to the results in Predator \& Prey, the MMM2 environment requires exploration, but learning is challenging with naive optimistic actions from risk-seeking policy. However, the ROE is able to effectively balance exploration-exploitation by adjusting the risk level over time in the more complex \textit{MMM2} environment, allowing for the identification of optimal strategies.

\textbf{Sensitiveness of Scheduling}
There should be proper exploration and exploitation steps to solve the complexity in RL environments. Here, we discuss the scheduling time of altering the region of quantile fractions, which has an impact on the exploration \& exploitation trade-off. Risk-seeking has the effect of exploration to search state or action, which results in a greater reward, but shifting the sampling region of distribution to the entire distribution shows exploitation based on the prior knowledge of distribution. As depicted in Figure \ref{fig:exploration_results} (mid, right),  the longer the scheduling period is, the higher the performance in \textit{Super hard} scenarios. In contrast, the longer the scheduling time in the \textit{Easy} scenario (Figure \ref{fig:exploration_results} left), which involves more exploitation than exploration, the worse the performance. Therefore, our dynamics of risk play a role in exploration and exploitation trade-offs, and appropriate risk scheduling steps are required.

\begin{figure}[!t]
  \centering
  \includegraphics[width=\linewidth]{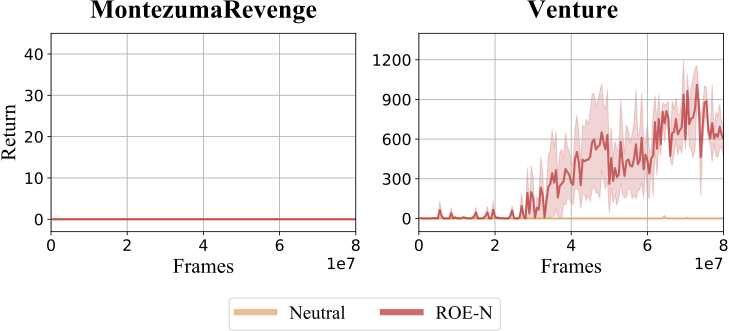}
  \caption{Episode return of IQN in Atari. The lines are the mean of 3 different risk-scheduling time steps with shaded areas representing a confidence interval of 25\% to 75\%.}
  \label{fig:atari_results}
  \Description{}
\end{figure}

\textbf{Without Aleatoric Uncertainty}
Although our method mainly focuses on addressing cooperative multi-agent environment, Figure \ref{fig:atari_results} shows our method's performance in a deterministic single-agent environment, Atari, which requires no cooperation. We conduct experiments on hard exploration games, Montezuma's Revenge, and Venture. In the early stages of training in sparse reward environments, such as Venture, our risk-based optimistic exploration displays excellent exploration. The reason for the improvements is a bit different but similar to other exploration approaches in distributional RL \cite{mavrin2019distributional, zhou2021non} with respect to considering the upper confidence of distribution. However, we need an intrinsic reward from distributional output to solve notably spare reward setting, such as Montezuma's Revenge.

\section{Conclusion \& Future work}
\label{sec:conclusion}
Endowing identical risk-seeking levels to agents makes them behave in a cooperatively optimistic manner.
Then shifting the sampling region of distribution to the entire distribution or lower region of distribution enables the agents to utilize the exploratory samples.
Experiments have demonstrated that ROE is effective in that it enhances the model's learning speed and improves final performance significantly more than other exploration techniques under aleatoric uncertainty for cooperative settings.
One important future work is to develop risk-based exploration for competitive environments.


\begin{acks}
This work was conducted by Center for Applied Research in Artificial Intelligence(CARAI) grant funded by Defense Acquisition Program Administration(DAPA) and Agency for Defense Development(ADD) (UD190031RD).
\end{acks}


\bibliographystyle{ACM-Reference-Format} 
\bibliography{mybibfile}

\onecolumn
\appendix
\begin{appendices}
\section{}
\label{sec:ablation}

\subsection{Hyperparameter}
We describe the hyperparameters that we utilized in multi-agent experiments in Table \ref{hyperparameter}. For the sake of fairness, the hyperparameters in the multi-agent are nearly identical to those in other works \cite{sun2021dfac, son2021disentangling}, with the exception of training steps. We anneal $\epsilon$ from 1.0 to 0.05 during the course of 50000 training steps and then fix for the remainder of the training duration. We fix $\gamma$ = 0.99. Replay buffer is capable of holding the most recent 5,000 episodes, and we randomly choose 32-size batches from the buffer. The target network is synchronized with the present network every 200 time steps. Each training has 8000000 steps. \ref{hyperparameter} and \ref{risk_scheduling_step} contains information about hyperparameters. In addition, we set to the default risk setting, environment-wise risk to be averse in DRIMA. We also describe the hyperparameters that we utilized in Atari experiments in Table \ref{hyperparameter_iqn}. We manually found the optimal hyperparameters for our experiments. $\epsilon$ is decaying linearly from 1.0 to 0.01, different with multi-agent experiments.

To schedule the risk in DMIX with ROE-A, we initialize $(\alpha, \beta) = (0.99, 1.0)$ in Equation \ref{eqn:scheduling}, and the same for ROE-N.
Similarly, $\alpha$ is linearly decayed to $0$, and when it does become $0$, $\beta$ begins to linearly decay to $0.25$.
Thus in the final phase, the sample range reaches and remains at $\mathcal{U}[0, 0.25]$ until the training is complete.
For DRIMA, the end state of the sampling region of quantile fractions is set to $\mathcal{U}[0, 0.1]$.
We performed a grid search for setting the number of scheduling steps as follows: \{10k, 25k, 50k\} for DMIX in SMAC, \{100k, 500k, 1M\} for DRIMA in SMAC, \{100k, 500k\} for DMIX in Predator \& Prey and \{800k, 2M, 4M\} for IQN in Atari using an environment- and algorithm-dependent risk-scheduling steps.

\begin{table}[!h]
\caption{Hyperparameters of SMAC and Predator \& Prey experiments}
\label{hyperparameter}
\centering
\begin{tabular}{c|cc p{0.5\linewidth}p{0.5\linewidth}}
\toprule
\textbf{Hyperparameter} & \textbf{Value} & \textbf{Description} \\ 
\midrule
Training steps  & 8000000, 800000 & how many steps was the model trained in SMAC and Predator \& Prey \\
Discount factor & 0.99 & how we estimate the future rewards \\
Learning rate & $5 \times 10^{-4}$ & learning rate by RMSProp optimizer \\
Target update period & 200 & update frequency of the target network \\
Replay buffer size & 5000 & prior samples' maximum container size \\
Batch size & 32 & quantity of samples per update \\
Batch size run & 5 & number of simultaneous simulators  \\
$\epsilon$ & 50000 & $\epsilon$-greedy exploration steps\\
Number of sampling $\tau$ & 8, 10 & number of quantile fraction samples in DFAC and DRIMA\\
\bottomrule
\end{tabular}
\end{table}

\begin{table}[!ht]
    \caption{Risk-scheduling steps reported on the results in SMAC}
    \label{risk_scheduling_step}
    \centering
    \resizebox{0.45\columnwidth}{!}{%
    \begin{tabular}{cccccc}
    \toprule
        & \multicolumn{2}{c}{\bf DMIX} & \multicolumn{2}{c}{\bf DRIMA} \\
    \cmidrule(r){2-3} \cmidrule(r){4-5} 
        & averse & neutral & averse & neutral\\
    \midrule
    \texttt{3s5z vs 3s6z} & 10,000 & 10,000 & 500,000 & 1000,000 \\
    \texttt{6h vs 8z} & 50,000 & 50,000 & 1,000,000 & 100,000 \\
    \texttt{corridor} & 50,000 & 50,000 & 1,000,000 & 1,000,000 \\
    \texttt{MMM2} & 10,000 & 10,000 & 1,000,000 & 1,000,000 \\
    \texttt{2s3z} & 50,000 & 10,000 & 100,000 & 100,000 \\
    \texttt{3s5z} & 50,000 & 50,000 & 1,000,000 & 500,000 \\
    \texttt{5m vs 6m} & 10,000 & 10,000 & 500,000 & 100,000 \\
    \texttt{3s vs 5z} & 10,000 & 10,000 & 1,000,000 & 1,000,000 \\
    \bottomrule
    \end{tabular}
    }%
\end{table}

\begin{table}[!h]
\caption{Hyperparameters of Single-agent experiments}
\label{hyperparameter_iqn}
\centering
\begin{tabular}{c|cc p{0.5\linewidth}p{0.5\linewidth}}
\toprule
\textbf{Hyperparameter} & \textbf{Value} & \textbf{Description} \\ 
\midrule
Frames  & 80000000 & how many frames was the model trained in Atari\\
Discount factor & 0.99 & how we estimate the future rewards \\
Learning rate & $5 \times 10^{-5}$ & learning rate by Adam optimizer \\
Target update period & 10000 & update frequency of the target network \\
Replay buffer size & 100000 & prior samples' maximum container size \\
Batch size & 32 & quantity of samples per update \\
Batch size run & 1 & number of simultaneous simulators  \\
$\epsilon$ & 250000 & $\epsilon$-greedy exploration steps\\
Number of sampling $\tau$ & 64 & number of quantile fraction samples in IQN\\
\bottomrule
\end{tabular}
\end{table}

\clearpage

\subsection{Additional Results}
In this section, we present Easy and Hard scenarios in SMAC. In addition, we report final performance in Table \ref{table:final_winrate}. We find that in Easy and Hard scenarios, win-rate converges more fast when using ROE than not using it.

\begin{table}[!h]
    \caption{Average win-rate (\%) performance of DMIX and DRIMA after training.}
    \label{table:final_winrate}
    \centering
    \resizebox{0.85\columnwidth}{!}{%
    \begin{tabular}{ccccccccc}
    \toprule
        & \multicolumn{4}{c}{\bf DMIX} & \multicolumn{4}{c}{\bf DRIMA} \\
    \cmidrule(r){2-5} \cmidrule(r){6-9}
        & \texttt{neutral} & \texttt{neutral$^{*}$} & \texttt{averse} & \texttt{averse$^{*}$} & \texttt{neutral} & \texttt{neutral$^{*}$} & \texttt{averse} & \texttt{averse$^{*}$} \\
    \midrule

3s5z vs 3s6z   & 0.6 \small $\pm$ 1.3 &  0.0 \small $\pm$ 0.0 &  1.3 \small $\pm$ 2.0 &  8.6 \small $\pm$ 8.6 & 0.0 \small $\pm$ 0.0 & 3.7 \small $\pm$ 3.3 & 1.8 \small $\pm$ 2.7 & 0.7 \small $\pm$ 1.3 \\
    
\cmidrule(r){1-9}
6h vs 8z   & 23.5 \small $\pm$ 17.1 &  34.0 \small $\pm$ 14.8 &  8.4 \small $\pm$ 7.2 &  12.4 \small $\pm$ 14.2 & 6.6 \small $\pm$ 5.8 & 21.1 \small $\pm$ 11.2 & 32.5 \small $\pm$ 7.8 & 38.5 \small $\pm$ 16.3  \\
                        
\cmidrule(r){1-9}
corridor & 53.3 \small $\pm$ 29.7 &  66.2 \small $\pm$ 22.6 &  12.2 \small $\pm$ 24.7 & 32.0 \small $\pm$ 38.9 & 54.0 \small $\pm$ 29.1 & 85.9 \small $\pm$ 5.1 & 92.2 \small $\pm$ 4.1 & 92.2 \small $\pm$ 5.2  \\
                        
\cmidrule(r){1-9}
MMM2  & 4.4 \small $\pm$ 8.3 &  51.8 \small $\pm$ 21.8 & 0.2 \small $\pm$ 0.9 & 60.7 \small $\pm$ 40.2 & 20.7 \small $\pm$ 14.7 & 30.3 \small $\pm$ 23.4 & 5.5 \small $\pm$ 10.6 & 54.4 \small $\pm$ 27.5  \\

\cmidrule(r){1-9}
2s3z  & 94.8 \small $\pm$ 4.6 &  94.8 \small $\pm$ 3.8 & 93.7 \small $\pm$ 6.0 & 99.1 \small $\pm$ 1.9 & 93.3 \small $\pm$ 4.7 & 94.0 \small $\pm$ 3.0 & 97.4 \small $\pm$ 2.6 & 92.9 \small $\pm$ 6.1  \\

\cmidrule(r){1-9}
3s5z  & 82.2 \small $\pm$ 14.8 &  84.9 \small $\pm$ 14.6 & 97.7 \small $\pm$ 3.9 & 98.2 \small $\pm$ 2.3 & 80.3 \small $\pm$ 10.9 & 86.6 \small $\pm$ 6.8 & 81.8 \small $\pm$ 8.0 & 91.4 \small $\pm$ 4.7  \\

\cmidrule(r){1-9}
5m vs 6m  & 60.7 \small $\pm$ 13.6 &  71.8 \small $\pm$ 7.5 & 69.6 \small $\pm$ 7.9 & 79.2 \small $\pm$ 6.9 & 77.4 \small $\pm$ 6.0 & 78.1 \small $\pm$ 5.9 & 71.1 \small $\pm$ 8.3 & 79.6 \small $\pm$ 6.1  \\

\cmidrule(r){1-9}
3s vs 5z  & 42.2 \small $\pm$ 26.9 &  60.4 \small $\pm$ 16.2 & 0.0 \small $\pm$ 0.0 & 0.0 \small $\pm$ 0.0 & 87.0 \small $\pm$ 8.2 & 93.3 \small $\pm$ 10.0 & 81.1 \small $\pm$ 11.5 & 87.7 \small $\pm$ 14.4  \\

    \bottomrule
    \multicolumn{9}{l}{\footnotesize $*$ : ROE} \\
    \end{tabular}
     }%
\end{table}

\begin{figure*}[h!]
    \centering
    \includegraphics[width=0.85\columnwidth]{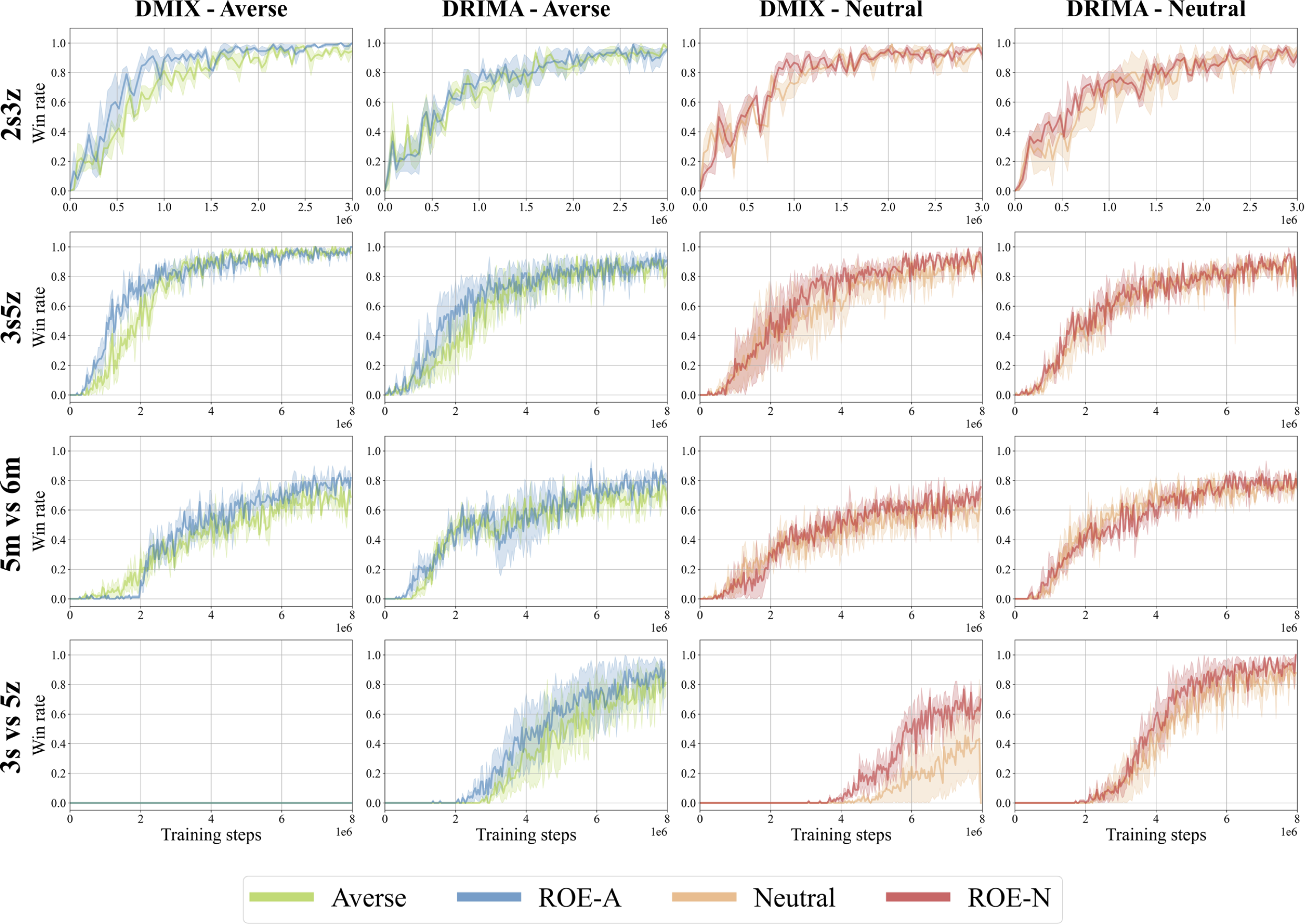}
    \centering
    \caption{Win-rate Results of DMIX and DRIMA in Easy, Hard scenarios. The label of X-axis and Y-axis represent algorithm - risk level and scenarios respectively. The lines are the mean of 3 random seeds using five parallel training with shaded areas representing a confidence interval of 25\% to 75\%.} 
    \label{fig:ablation_results}
\end{figure*}

\clearpage

\subsection{Behaviors in SMAC}
\begin{figure*}[h!]
    \centering
    \includegraphics[width=\columnwidth]{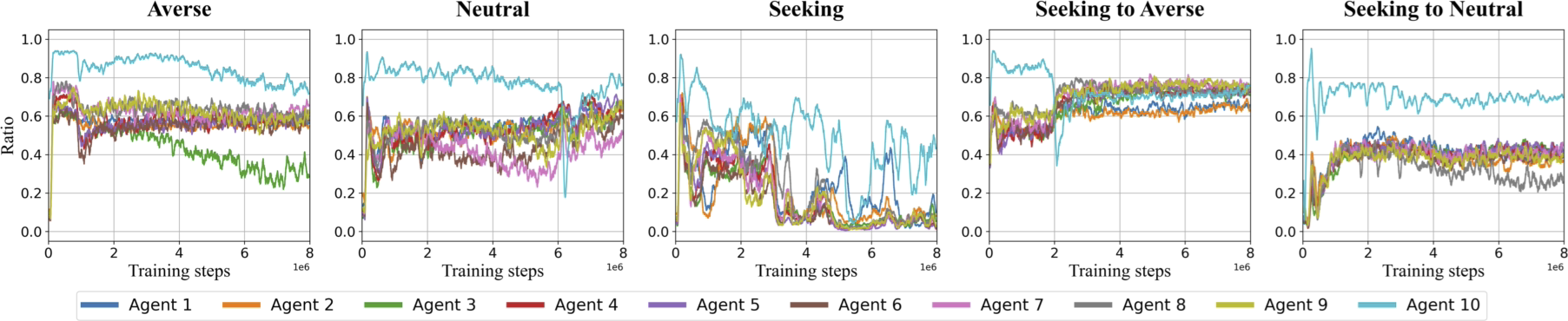}
        \centering
    \caption{The movement-attack ratio of DMIX in \texttt{MMM2} in SMAC. Seeking to averse, Agent 10 is healer which have a different role in this scenario. We select the seed with the highest win rate from 5 random seeds for comparison.} 
    \label{fig:ablation_behavior_results}
\end{figure*}

We conduct ablation experiments to demonstrate how our method works in the SMAC \texttt{MMM2} scenario. Since we desire to figure out how ROE works, we set the scheduling step to 100k, which is longer than our search space in SMAC experiments. The win rates of seeking to averse and seeking to neutral agents are 0.866 and 0.75, respectively. In Figure \ref{fig:ablation_behavior_results}, the movement-attack ratio is used to compare risk-averse, neutral, seeking, seeking to averse, and seeking to neutral. The movement-attack ratio is the proportion of actions within a single episode. The closer to 1, the more attacking(for agent 10, a healer, healing), and the closer to 0, the more the movement is. Initially, we could observe that the method of training is distinct. In the early stages of training, agents with a static risk level concentrate on the attack before attempting to find the other strategy. In contrast, ROE agents behave differently. They initiate with actions other than attacks, and then they deal damage to enemies. It is comparable to the process of exploration and exploitation. They determine the winning strategy first, then employ it. We hypothesize that these distinctions make our method more effective than other risk levels.

\subsection{Algorithm}
\paragraph{DFAC} Based on the IQN algorithm \cite{dabney2018implicit}, DFAC \citep{sun2021dfac} is the first approach to integrate distributional RL and multi-agent RL. For distributional output sampling quantile fractions from $\mathcal{U}[0, 1]$ and approximating return values with quantile regression, IQN was utilized in DFAC. By utilizing mean-shape decomposition, the authors successfully incorporated distributional viewpoint into a multi-agent framework without violating the IGM criterion, which may be expressed as follows:
\begin{equation}
    \begin{aligned}
    \label{dfac_igm}
        \mathrm{arg\,max}_{\textbf{a}}\mathbb{E}[Z_{joint}(\textbf{x, a})] = \begin{pmatrix}
        \mathrm{arg\,max_{a_{1}}} \mathbb{E}[Z_{1}(\mathrm{x_{1}, a_{1}})]\\
        \vdots \\
        \mathrm{arg\,max_{a_{N}}} \mathbb{E}[Z_{N}(\mathrm{x_{N}, a_{N}})]\\
        \end{pmatrix}    
    \end{aligned}
\end{equation}
that can be proved by the DFAC Theorem which is proven to meet the IGM condition. DFAC outperforms all other algorithms, especially in difficult scenarios. This approach may also be modified to work with IQL, VDN\cite{sunehag2017value}, and QMIX\cite{rashid2018qmix}. The DMIX variations of the DFAC algorithm, which combines with QMIX, are employed as our baseline.

\paragraph{DRIMA} DFAC only examines a single risk source, however DRIMA \cite{son2021disentangling} considers splitting risk sources into agent-wise risk $w_{agt}$ and environment-wise risk $w_{env}$, creating an additional hyperparameter in contrast to the DFAC algorithm's agent-wise risk hyperparameter. Environment-wise risk may be regarded as transition stochasticity, whereas agent-wise risk is the unpredictability caused by the actions of other agents that cannot be represented by environment MDP (Markov Decision Process). In distributional multi-agent reinforcement learning methods, models use the risk level as an input to the agent utility function, which outputs a distribution of return per action that can be interpreted by agents as randomness. In contrast, in DRIMA, the agent receives agent-specific risk, and in the process of determining the joint distribution of returns, the joint action-value network serves as an input, agent utility function and joint action-value network having a hierarchical architecture resembling the structure of QTRAN \cite{son2019qtran}. Agent-wise utility function, true action-value network, and transformed action-value network comprise DRIMA's network architecture. Deep Recurrent Network takes $w_{agt}$ as an input and structures an agent-specific utility function. True action-value network approximates the true distribution of returns with extra representation power that gets the environment-wise risk $w_{env}$, the state $x$, and the outputs of utility functions $\mathbb{Z}_{i}$.


\clearpage
\end{appendices}


  




\end{document}